\documentclass[conference]{IEEEtran}

\usepackage{graphicx}
\usepackage{float}
\usepackage[cmex10]{amsmath} 
\usepackage{multirow}
\usepackage{hhline}
\usepackage{mwe} 
\usepackage{subcaption}
\usepackage{cite}
\usepackage[numbers,sort&compress]{natbib}
\usepackage{tabularx}
\usepackage{lipsum,booktabs,siunitx}
\newcolumntype{T}[1]{S[table-format=#1,group-digits=false]}
\usepackage{xcolor}
\usepackage{colortbl}
\usepackage{lipsum}
\usepackage{booktabs}

\usepackage{array}
\usepackage{caption}
\usepackage{url}

\begin{document}
\bstctlcite{IEEEexample:BSTcontrol}
\title{High Order Local Directional Pattern Based Pyramidal Multi-structure for Robust Face Recognition
}

\author{\IEEEauthorblockN{Almabrok Essa\IEEEauthorrefmark{1} and Vijayan Asari\IEEEauthorrefmark{2}}
	\IEEEauthorblockA{\IEEEauthorrefmark{1}Department of Electrical Engineering and Computer Science, Cleveland State University, Cleveland, {OH}, USA\\
	\IEEEauthorrefmark{2}Department of Electrical and Computer Engineering, University of Dayton, Dayton, {OH}, USA\\
	Email:\IEEEauthorrefmark{1}a.essa@csuohio.edu; \IEEEauthorrefmark{2}vasari1@udayton.edu}}
\maketitle

\begin{abstract}
Derived from a general definition of texture in a local neighborhood, local directional pattern (LDP) encodes the directional information in the small local $3\times3$ neighborhood of a pixel, which may fail to extract detailed information especially during changes in the input image due to illumination variations. Therefore, in this paper we introduce a novel feature extraction technique that calculates the $n^{th}$ order direction variation patterns, named high order local directional pattern (HOLDP). The proposed HOLDP can capture more detailed discriminative information than the conventional LDP. Unlike the LDP operator, our proposed technique extracts $n^{th}$-order local information by encoding various distinctive spatial relationships from each neighborhood layer of a pixel in the pyramidal multi-structure way. Then we concatenate the feature vector of each neighborhood layer to form the final HOLDP feature vector. The performance evaluation of the proposed HOLDP algorithm is conducted on several publicly available face databases and observed the superiority of HOLDP under extreme illumination conditions.

\end{abstract} 


\section{Introduction}

Face recognition has received a great deal of attention in recent years and become an active research topic in the fields of computer vision, image processing, pattern recognition, and machine learning. Face recognition has spread in several applications such as biometric systems, access control and information security systems, surveillance systems, content-based video retrieval systems, credit-card verification systems, and more generally image understanding. The key of each face recognition system is finding an efficient and discriminative feature extraction technique that must be able to extract features from the face image, which are distinct and stable under different conditions during the image acquisition process, like illumination variation, random noise, and alignment error which have negative influences on the detection and recognition accuracy \cite{1}. In the context of feature description and representation applications, there are two common types of techniques which are subspace based holistic features and local appearance based features.

The most successful local appearance based feature algorithms group is based on the concept of a spatial histogram model local pattern descriptors that includes local binary pattern (LBP) \cite{4}, local ternary pattern (LTP) \cite{6}, local edge/corner feature integration (LFI) \cite{essa4}, local directional pattern (LDP) \cite{5}, and local boosted features (LBF) \cite{essa6}. Also, there are some methods that combine theses two different types of techniques like \cite{essa5}. These descriptors usually have been used in the field of face recognition and facial expression recognition for still images, since local pattern descriptors have quite important properties to be robust against uncontrolled environments such as illumination variation, random noise, and alignment error, as well as the computational simplicity. In addition, some of these kind of descriptors could be modified to apply for videos as in \cite{essa3,essa7}. The main goal of the local pattern descriptors is extracting the image features that are distinct and stable under different conditions during the image acquisition process. 

The original LBP operator was introduced by Ojala et al. \cite{4} for texture analysis, and has proved a simple yet powerful approach to describe local structures. LBP was originally defined for $3\times3$ neighborhood pixels, which gives an $8$ bit binary code that is derived from comparing each pixel with its central pixel. If a neighbor pixel has a higher intensity value than the center pixel (or the same intensity value) then a $1$ is assigned to that pixel, which is otherwise a $0$. Formally, it takes the form 

\begin{equation}
LBP = \sum_{p=0}^{7}f(d_{p}-d_{c})\times 2^{p}
\end{equation}
and
\begin{equation}\label{eq:fx}
f(x) = 
\begin{cases} 
1 & \text{if } x \geq 0 \\
0 & \text{if } x < 0
\end{cases}
\end{equation}
\noindent
where $d_{c}$ and $d_{p}$ denote the intensity values of the central pixel and its surrounding pixels respectively. 

LBP operator has a number of extensions that have been extensively used in many applications such as, face image analysis \cite{7,8}, image and video retrieval \cite{9,10}, biomedical and aerial image analysis \cite{11,12}. Some of the extended versions of LBP were using neighborhoods of different sizes and rotation invariant, which allow to deal with large scale structures that may be the best representative features of some types of textures as well as detect the uniform patterns for any quantization of the angular space \cite{13,14}. Moreover, a single LBP code can be generated for each pixel in an image using all the previously mentioned extensions. Then to collect the occurrences of different LBP codes from all pixels, a histogram is created. This causes two fundamental limitations, trade-off between resolution and the presence of noise. To overcome that, a Fuzzy Local Binary Pattern (FLBP) is introduced in \cite{flbp,15}. FLBP extends the LBP by incorporating fuzzy logic in the representation of local patterns of texture. Fuzzification allows FLBP to contribute to more than a single bin in the distribution of the LBP values used as a feature vector.           

Another limitation of LBP is that the original LBP operator does not consider the texture features from the magnitude component of the image local differences, as well as the local features from multi-resolution of the image, so that a completed LBP (CLBP) has been introduced in \cite{clbp}. CLBP represents a local region of an image by its local central information and a local difference sign-magnitude transform (LDSMT). One of the most generalization of the LBP is local ternary pattern (LTP) \cite{ltp}. LTP is more discriminant and less sensitive to noise in uniform regions than LBP. Unlike the LBP operator, LTP extends the LBP to $3$ codes, in which intensity value in a tolerance interval zone of width $\pm \tau$ around the center pixel is assigned to $0$. If a neighbor pixel has a higher intensity value than this zone then a $1$ is assigned to that pixel, which is $-1$ if it is below this.             

\begin{figure}[h!]
	\captionsetup{skip=12pt}
	\centering
	\includegraphics[width=.47\textwidth]{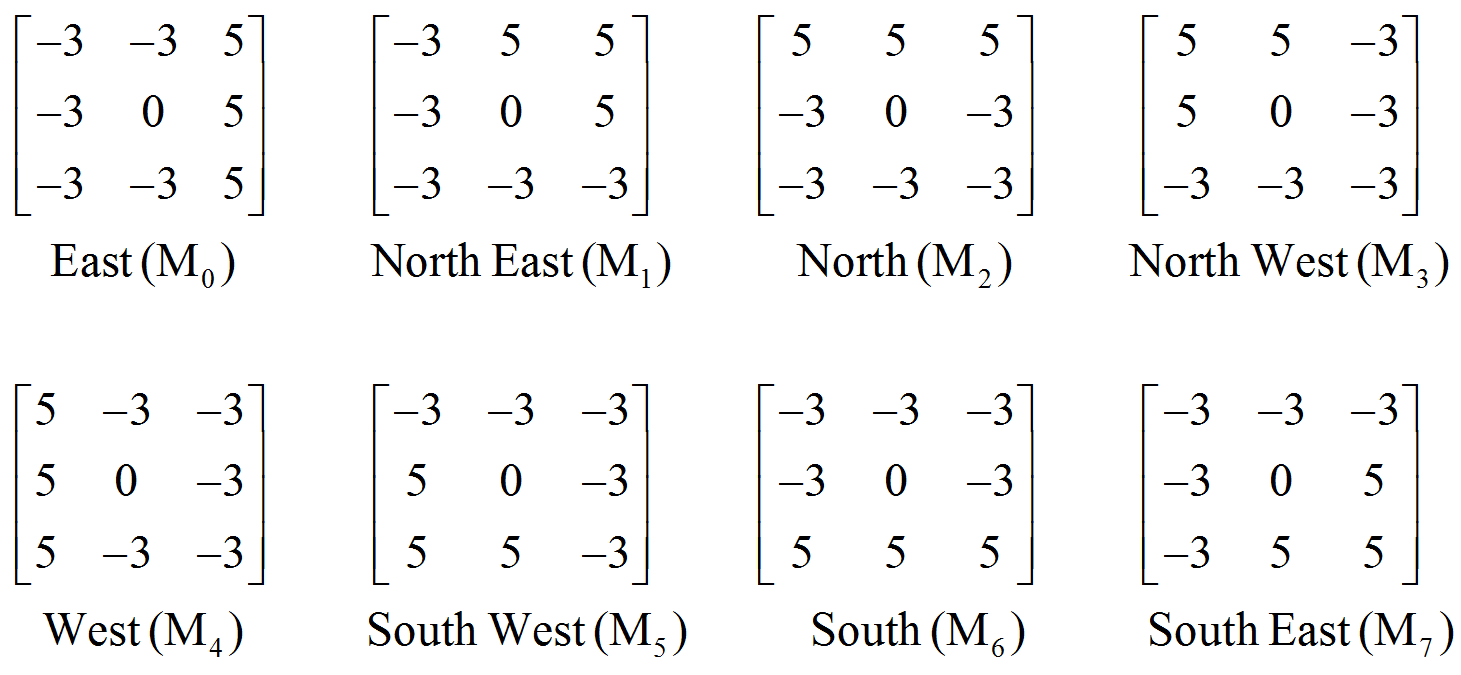}
	\caption{Kirsch edge masks in all eight directions}
	\label{fig-2-2}
\end{figure}

\begin{figure}[h!]
	\captionsetup{skip=15pt}
	\centering
	\includegraphics[width=.46\textwidth]{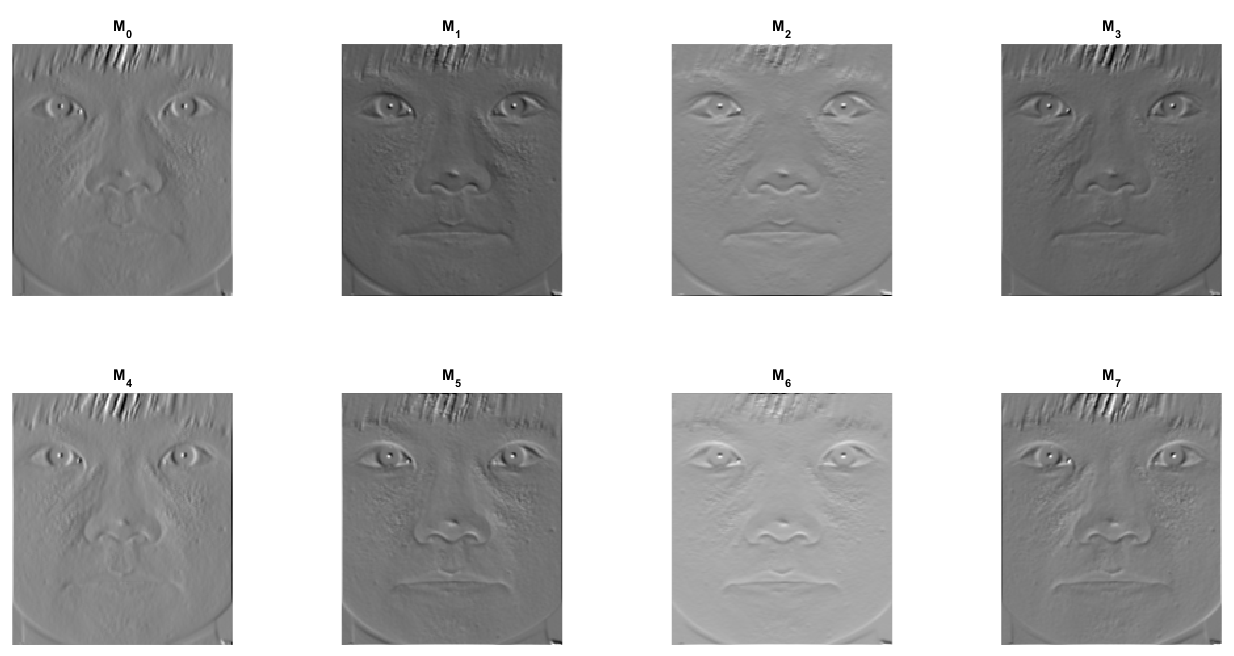}
	\caption{Kirsch kernel filtered output images}
	\label{fig-4-3}       
\end{figure}

One of the newest local appearance methods that has been tried to avoid the shortcoming of LBP is local directional pattern (LDP). LDP encodes the directional information in the neighborhood instead of the direct intensity as LBP does with higher computational cost. LDP computes the edge response values in eight different directions at each pixel position, and uses the relative strength magnitude to encode the image texture, by the help of Kirsch compass gradient masks in eight different directions \cite{5}. Kirsch mask is a first derivate filter which is used to detect edges in all eight directions of a compass considering all eight neighbors \cite{Kirsch}. Specifically, it takes a single mask, denoted as $M_i(x,y)$ for $i = 0,1,...,7$, and rotates it in $45^{\circ}$ increments through all 8 compass directions as shown in  Fig. \ref{fig-2-2}. An example of Kirsch kernel filtered images for one sample face image can be seen in Fig. \ref{fig-4-3}, where all eight directional features are extracted with their corresponding masks.

The presence of the edge is determined by the mask that produces the maximum output value at certain orientation. Therefore, we need to know the $t$ most prominent directions to generate the LDP code. Then, the top $t$ directional bit responses are set to 1 and the other (8-t) bits of the 8-bit LDP pattern are set to 0. Formally, the LDP code can be derived by 

\begin{equation}
LDP = \sum_{i=0}^{7}f(m_{i}-m_{t})\times 2^{i}  
\end{equation}
\noindent
where $m_{t}$ is the $t^{th}$ most significant directional responses and $m_{i}$ denotes the intensity values of the neighbor pixels for $i=0,1,...,7$. The thresholding function $f(x)$ can be defined as in Eq. \eqref{eq:fx}.

LDP has been used extensively in face and facial expression recognition \cite{23,24,25,26}. Since the edge responses are less sensitive to noise and illumination than the intensity values, the resultant of using the edge responses to extract the image features is more stable and maintains more information than using the intensity values. In addition, describing the local primitives including different types of curves, corners, and junctions could be more stable. 

Nevertheless, both above mentioned descriptors LBP and LDP consider only the first order intensity pattern change in a local neighborhood, which may fail to extract detailed information especially during changes in face image due to the noise and illumination variation problems. 

In this paper, we introduce a new feature extraction technique to calculate the $n^{th}$ order directional patterns, named High Order Local Directional Pattern (HOLDP). After that, the spatial histogram is described for modeling the distribution of HOLDP of a face image. Lastly, several experimental results and comparison with a set of the state-of-the-art methods will be given to show the strength of HOLDP for face recognition task under extreme illumination conditions.

The rest of the paper is organized as follows. Section 2 provides the mathematical details of the proposed HOLDP technique. The discussion on the datasets and experimental results are presented in Section 3. Finally, the conclusion is drawn in Section 4.

\section{The Proposed Scheme}
In this correspondence, a high order local directional pattern descriptor is proposed for robust face recognition under extreme illumination and lighting variations. The conventional LDP encodes the directional information in the small local neighborhood of a pixel ($3\times3$) window size, which may fail to extract detailed information especially at the presence of uncontrolled environments during the image acquisition process such as random noise and illumination variation problems, etc. The proposed HOLDP can capture more detailed discriminative information than the LDP. Unlike the LDP operator, our proposed technique extracts $n^{th}$ order local information by encoding various distinctive spatial relationships from each neighborhood layer of a pixel with the pyramidal multi-structure way. Then we concatenate the feature vector of each neighborhood layer to form the final HOLDP feature vector. Several observations can be made for the proposed technique HOLDP:

\begin{itemize}
	\item Under the proposed framework, the well known LDP is a special case of HOLDP, which simply calculates the $1^{st}$ order pattern information in the local neighborhood of a pixel.
	\item The relation between the neighbor layers and the pixel under consideration could be easily weighted in HOLDP based on the distance between each layer and the central pixel. Because of that, the pixels within the closest layer to the central pixel have more weight than the others.
	\item Due to the same format and feature length of different order HOLDPs, they can be readily fused, and the accuracy of the face recognition can be significantly improved after the fusion. 	
\end{itemize}

\subsection{Theoretical Analysis}

\begin{figure}[h!]
	\captionsetup{skip=10pt}
	\centering
	\includegraphics[width=.45\textwidth]{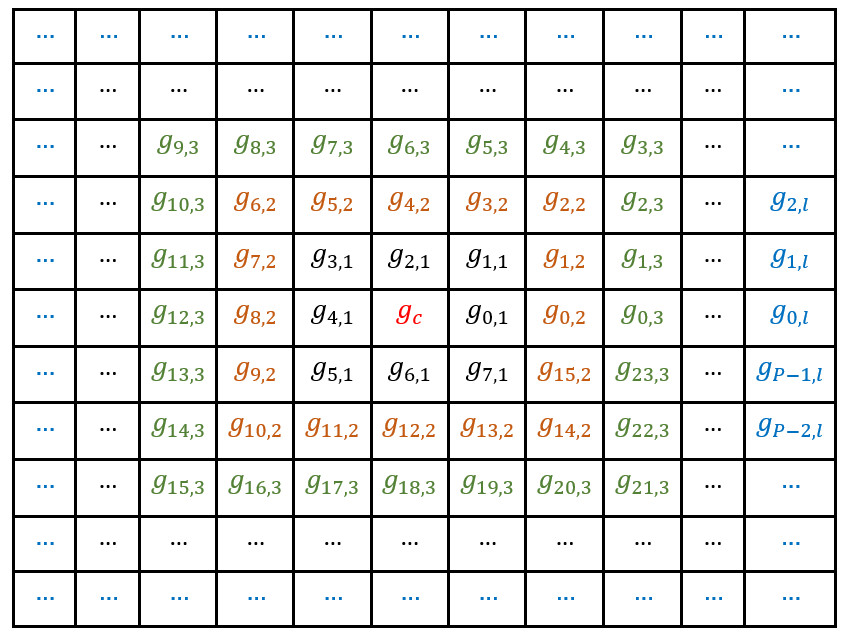}
	\caption{Central pixel and its $P$ evenly spaced neighbors within $l$ neighboring layers.}
	\label{fig-4-1}
\end{figure}

Given a central pixel $g_{c}$ in the image and its $P$ evenly spaced $l$ neighbors $g_{p,l}$ where $p = 0,1, ..., P-1$ and $l = 1,2,..., n$ (number of neighboring layers) as can be seen in Fig. \ref{fig-4-1}. The traditional LDP simply calculate the $1^{st}$ order edge directional value along $p$ direction, with limitations of the total number of involved neighbors $P=8$ and $l=1$. To provide a stronger discriminative capability in describing detailed texture information than the $1^{st}$ order as used in LDP, we propose to use different layers of neighborhood configuration. 


The directional edge extraction process can be done by convolving the input image $I(x,y)$ with Kirsch masks in eight different directions $M_i(x,y)$ for $i = 0,1,...,7$, which can be seen in Fig. \ref{fig-2-2}.

\begin{figure}[h!]
	\centering
	\includegraphics[width=.5\textwidth]{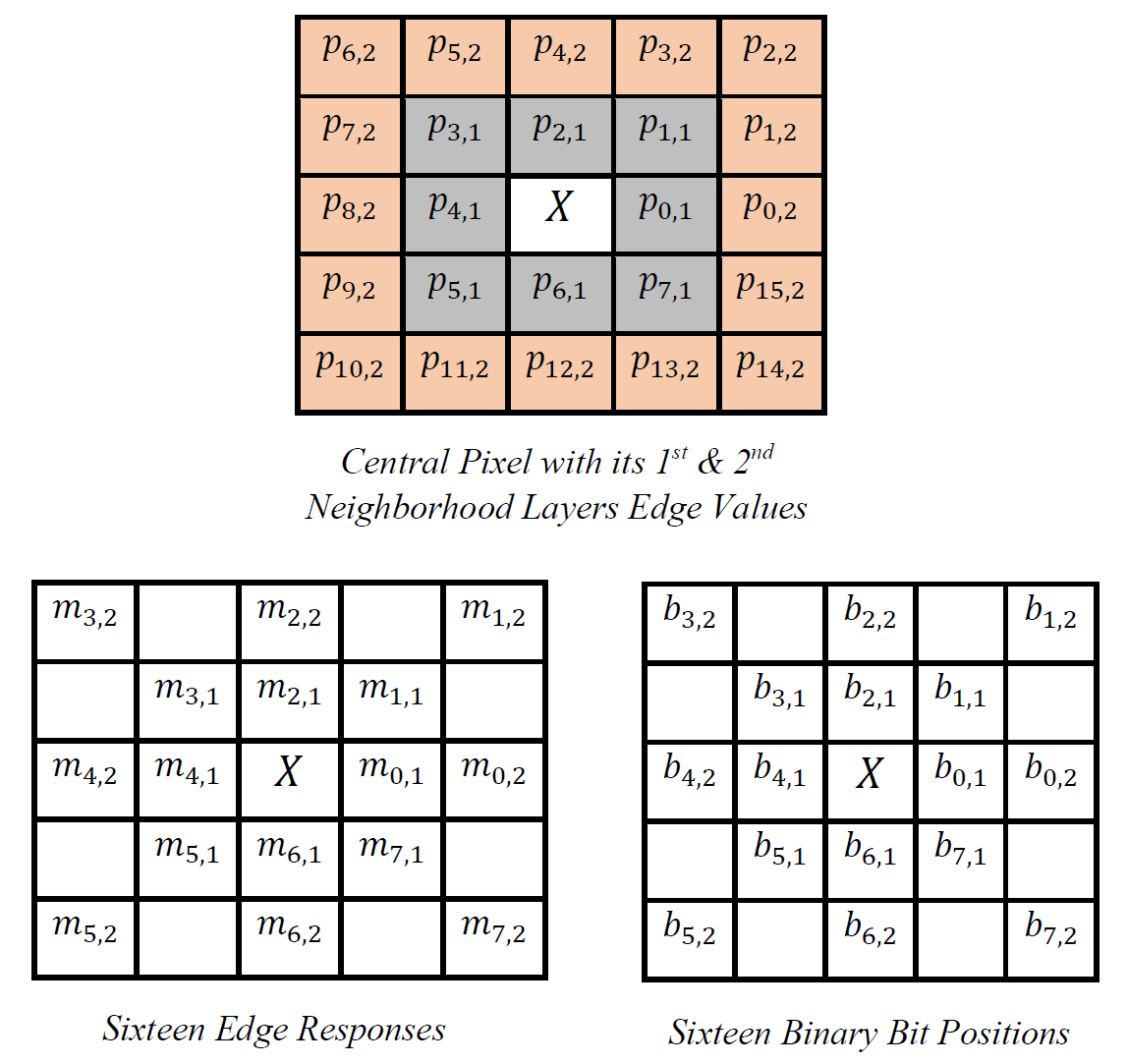}
	\caption{Two neighborhood layers and sixteen edge responses with their binary bit positions.}
	\label{fig-4-4}       
\end{figure}

To make our calculation simple and easy to compute the high order relevant edge values along $p$ directions, let us assume $P=8$ and $l=1$ to calculate the $1^{st}$ order, which means ($3\times3$) neighborhood pixels, $P=16$ and $l=2$ to calculate the $2^{nd}$ order for ($5\times5$) neighborhood. That means twenty four neighbor pixels will be under consideration, as can be seen in Fig. \ref{fig-4-4} top, and $P=8 \times n$ and $l=n$ to calculate the $n^{th}$ order. 

For example, the $1^{st}$ and $2^{nd}$ order local directional patterns $(HOLDP_{1})$ and $(HOLDP_{2})$ could be computed as follows; Firstly, we need to find the eight different directional edge response values of the first and second neighborhood layers ($1^{st}$ and $2^{nd}$ LDP order), which can be done by 

\begin{equation}
m_{i,1} = p_{i,1}, 
\end{equation}
\noindent
\setlength{\jot}{30pt}
\begin{equation} \label{1}
m_{i,2} = \frac{1}{3} \sum_{j=-1}^{j=1} p_{g,2},  
\end{equation}
\noindent 
and
\begin{equation} \label{2}
g = \mod(2i+j,16), \qquad \ for \ i = 0,1,...,7 
\end{equation}

\noindent 
where $m_{i,1}$ and $m_{i,2}$ are the eight different directional edge response values of the first and second neighborhood layers respectively. $p_{g,2}$ is the edge response value after convolving the input image with Kirsch kernels, the subscripts $g$ and $2$ are the number of surrounding pixels of each direction $i$ and the second neighborhood layer (second order) respectively, and $(mod)$ is the modulo operation.

From equations~\eqref{1} and~\eqref{2}, two observations can be made. The first one is that by using the modulo operation $(mod)$, we maintain the circular neighbors configuration. The second one is that by taking the mean of each three pixels in this case ($2^{nd}$ order), we give less weight to the pixels in the second layer than the pixels in the first layer, which reduces the total number of second layer's pixels from sixteen pixels to eight. For example, the edge response value at $0^{\circ}$ direction can be calculated as 

\begin{equation}
m_{0,2} = \frac{p_{15,2}+p_{0,2}+p_{1,2}}{3}
\end{equation}

Secondly, based on the observation that every corner or edge has high response values in particular directions, we are interested to know $t$ the most prominent directions after convolving the input image with all $8$ masks. Then the local directional pattern of each pixel position in each neighbor layer for the $1^{st}$ and $2^{nd}$ order can be formed as

\begin{equation}
HOLDP_{1} = \sum_{i=0}^{7} f( m_{i,1} - m_{t,1}) \times 2^{i}   
\end{equation}
\noindent
and
\begin{equation} 
HOLDP_{2} = \sum_{i=0}^{7} f(m_{i,2} - m_{t,2}) \times 2^{i}
\end{equation}

\noindent
where $m_{t,1}$ and $m_{t,2}$ are the $t^{th}$ most significant directional responses of the first and second neighboring layers respectively. The thresholding function $f(x)$ can be defined as in Eq. \eqref{eq:fx}.  

After identifying the local directional pattern of each pixel in each neighborhood layer ($HOLDP_{1}$ for the first layer and $HOLDP_{2}$ for the second layer), a histogram is built to represent the whole distinguishing features of an image from each neighbor layer separately. Then we concatenate these two histograms to finalize the feature vector of our proposed technique, which statistically describes the face image characteristics. This way, it is easy to combine multiple HOLDP with more than two orders, by concatenating different histograms one by one. 

The $3^{rd}$ order local directional pattern $(HOLDP_{3})$ could be computed by concatenating the $2^{nd}$ order local directional pattern histogram that has been obtained above with the local directional pattern histogram of the third neighborhood layer, which can be found by

\begin{equation} 
HOLDP_{3} = \sum_{i=0}^{7} f(m_{i,3} - m_{t,3}) \times 2^{i}
\end{equation}

\noindent
where the thresholding function $f(x)$ can be defined as in Eq. \eqref{eq:fx}. $m_{t,3}$ is the $t^{th}$ most significant directional responses and $m_{i,3}$ is the eight different directional edge response values of the third neighborhood layer, which is computed by

\begin{equation} \label{third}
m_{i,3} = \frac{1}{5} \sum_{j=-2}^{j=2} p_{g,3} 
\end{equation}
\noindent
and
\begin{equation}
g = \mod(3i+j,24), \qquad \ for \ i={0,1,...,7}  
\end{equation}

\begin{figure}[b!]
	\centering
	\includegraphics[width=.485\textwidth]{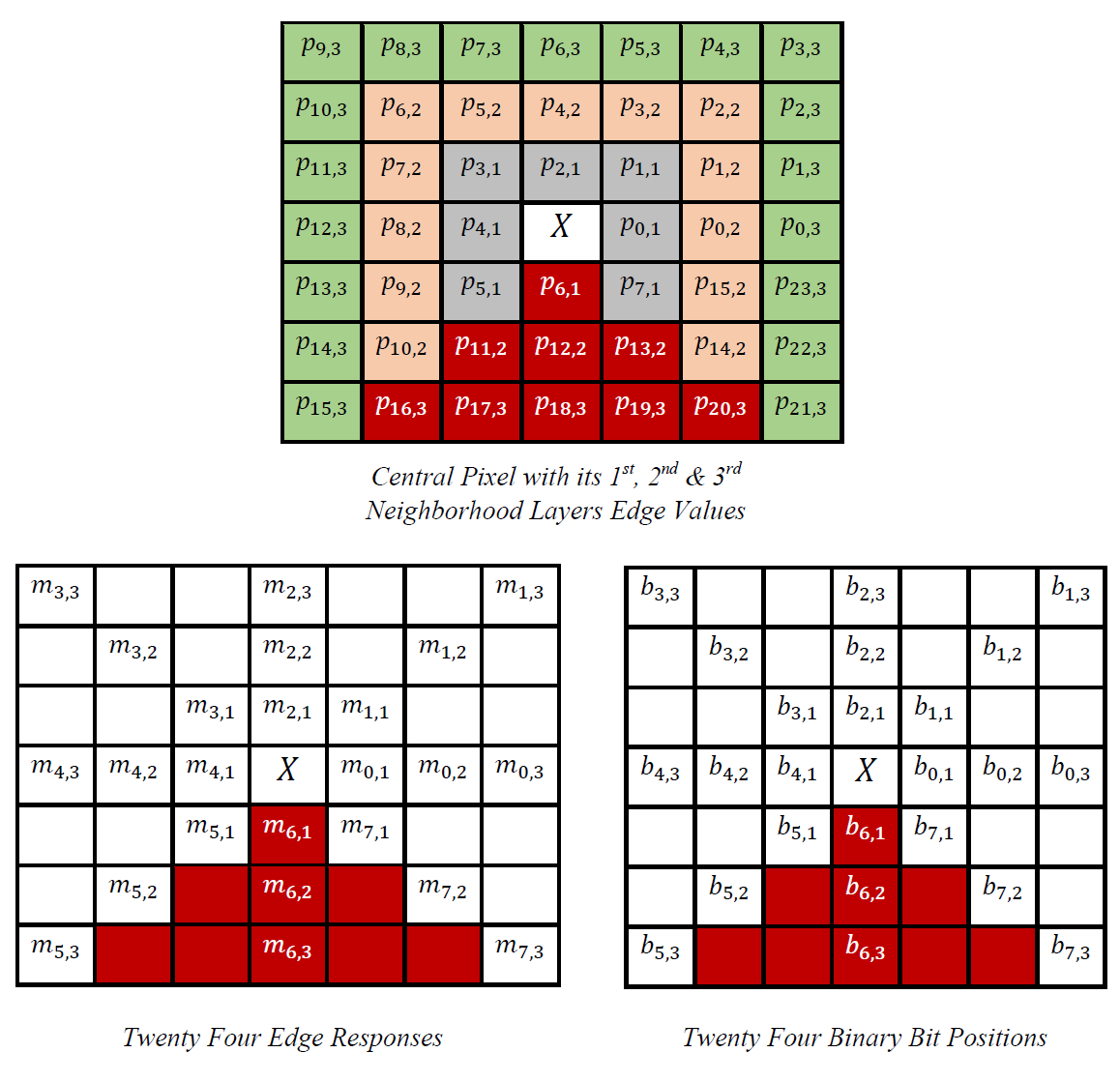}
	\caption{Three neighborhood layers and twenty four edge responses with their binary bit positions.}
	\label{fig-4-5}       
\end{figure}

From equation~\eqref{third}, we can say that using the pyramidal multi-structure approach, which can be seen in Fig. \ref{fig-4-5} (the red cells), we give less weight to the pixels in the third layer than the pixels in the second layer, which is less than the ones in the first layer. In other words, to get the edge response value for each direction, we consider five pixels as one pixel from the third layer, three pixels as one pixel from the second layer, and one pixel from the first layer. For example, the edge response values at $270^{\circ}$ direction using the pyramidal multi-structure approach can be calculated as 
\begin{equation}
m_{6,1} = p_{6,1} \ ,
\end{equation}
\setlength{\jot}{5pt}
\begin{equation}
m_{6,2} = \frac{p_{11,2}+p_{12,2}+p_{13,2}}{3} ,  
\end{equation}
and
\begin{equation}
m_{6,3} = \frac{p_{16,3}+p_{17,3}+p_{18,3}+p_{19,3}+p_{20,3}}{5}
\end{equation}

Based to the analysis above, the $n^{th}$ order local directional pattern $(HOLDP_{n})$ of each pixel position in each neighbor layer from the input image can be defined as

\begin{equation}\label{fourth}
HOLDP_{n} = \sum_{i=0}^{7} f(m_{i,n} - m_{t,n}) \times 2^{i} 
\end{equation}
\noindent
where $n=1,2,...$ is the local directional pattern order (the number of neighborhood layers), $m_{t,n}$ is the $t^{th}$ most significant directional responses of each neighboring layer $n$, the thresholding function $f(x)$ can be defined as in Eq. \eqref{eq:fx}, and $m_{i,n}$ is the eight different directional edge response values of each neighborhood layer $n$, which can be computed as

\begin{equation}\label{fifth}
m_{i,n} = \frac{1}{2n-1} \sum_{j=-n+1}^{n-1} p_{g,n} 
\end{equation}
and 
\begin{equation}\label{sixth}
g = \mod(ni+j,8n) \qquad \ for \ i={0,1,...,7}
\end{equation}

\subsection{The Adaptiveness of the Proposed Technique}
After the eight different directional edge response values of each neighboring layer are computed, the image encoding and decoding (IED) strategy is applied \cite{essa1, essa2}. Unlike the conventional LDP which needs to know the most prominent edge values to set them to $1$ and the rest to $0$, the local directional patterns in this section can be formed adaptively by comparing the $8$ neighboring pixels (excluding the central pixel) with their median of each neighborhood layer. If a neighboring pixel has a higher edge value than the median value (or the same value) then a $1$ is assigned to that pixel, which is otherwise a $0$. We tried to use other different measures to adaptively find the relationship of the neighboring pixels of each layer such as thresholding the pixels with their mean, standard deviation, and variance, but we found that the median provides better accuracy rates. Therefore, the $n^{th}$ order adaptive local directional pattern $(AHOLDP_{n})$ of each pixel position in each neighbor layer from the input image can be defined as
\begin{equation}
AHOLDP_{n} = \sum_{i=0}^{7} f(m_{i,n} - m_{n}) \times 2^{i} 
\end{equation}
\noindent
where $n=1,2,...$ is the local directional pattern order (the number of neighborhood layers), $m_{n}$ is the median of each neighboring layer $n$, and $m_{i,n}$ for $i = 0,1,...,7$ is the eight different directional edge response values of each neighborhood layer $n$, which can be computed using equations~\ref{fifth} and~\ref{sixth}. The thresholding function $f(x)$ can be defined as in Eq. \eqref{eq:fx}.

\section{Experimental Results}

For evaluation, we used four face recognition benchmarks that include diversity of illumination and lighting conditions, namely extended Yale B database \cite{yaleb1,yaleb2}, AT\&T (ORL) dataset \cite{orl}, Georgia Tech (GT) face database \cite{gt}, and AR database \cite{ar1,ar2}. All the images were resized to $64\times64$. After that, we extracted the information from each image using our proposed technique HOLDP and represented it as one histogram vector. The length of this feature vector (histogram) depends on the order of the local directional pattern descriptor, which means it is $n\times 256$. For example, the first order is $256$ bins, the second order is $512$ bins, and so on. In addition, the proposed HOLDP and AHOLDP techniques are compared with five spatial feature extraction methods that have common characteristics including LBP, CLBP, FLBP, LTP, and the conventional LDP, which considers as a special case of the proposed HOLDP technique (the $1^{st}$ order). When it comes to the face recognition process, the objective is to compare the encoded feature vector from one image with all other candidate feature vectors using a library for support vector machines classifier (LIBSVM) \cite{libsvm}.

Two different experiments are conducted to verify the effectiveness and efficiency of the proposed HOLDP framework. The first one is exploring the effectiveness of different local directional order (different neighborhood layers for each pixel of the image) as changing the number of the most prominent response values $\{t=2,3,\cdots,6\}$. The second one is evaluating the effectiveness of the proposed HOLDP and AHOLDP by comparing them with five popular texture feature extractors including LBP, CLBP, FLBP, LTP and LDP. To avoid any bias, we randomly selected the data for training and testing, then the experiments were repeated $10$ times, after that the average result is calculated for comparison. Note that, we coded LDP technique since there is no source code publicly available and it is a special case of our proposed HOLDP technique. 


\subsection{Extended Yale B Database}
\begin{figure}[!h]
	\centering
	\hspace*{-0.2 cm}
	\includegraphics[width=.47\textwidth]{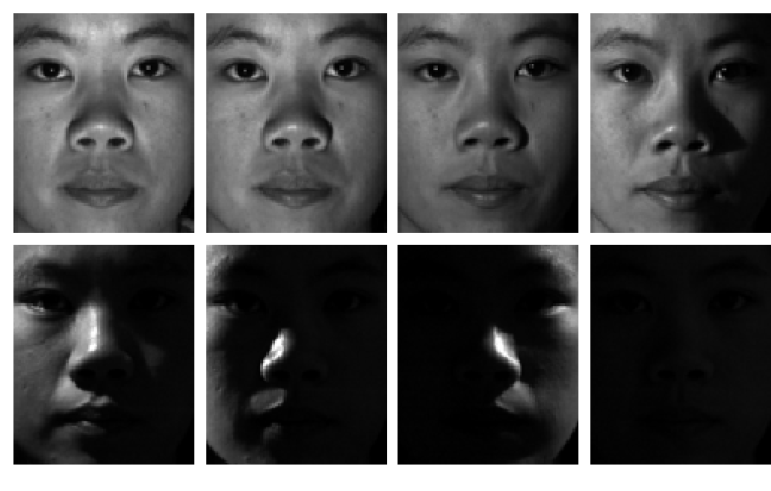}
	\caption{Samples of one subject from the extended Yale B database.}
	\label{fig-7}       
\end{figure}

The extended Yale B database has a total of $2280$ face images of $38$ subjects representing $60$ illumination conditions per subject under the frontal pose, Fig. \ref{fig-7} shows some sample faces of one subject of this dataset. In the figure, it is clear how the illumination problems extremely affect the input images. To show the effectiveness of the proposed HOLDP technique, we randomly selected half of the data for training ($30$ images/subject) and the other half for testing. Then we summarized the highest recognition rates as changing the number of neighborhood layers (the order of the proposed approach) in the range $(1-4)$ along with changing the threshold $t$ (the most prominent edge response values) in the range $\{t=2,3,\cdots,6\}$ as well as with the adaptive HOLDP for each pixel of the input image in Table \ref{table-1}. 

The performance results of well known local appearance based feature algorithms for face recognition  like LBP, CLBP, FLBP, LTP, and LDP with the proposed methods HOLDP and AHOLDP on extended Yale B dataset are presented in Table \ref{table-2}. Note that we ran the same sets of training and testing for all methods, since LBP, CLBP, and FLBP are nonparametric methods, and we extract their histograms directly. For LTP approach, there is one main parameter $\tau$ that splits LTP into positive and negative parts, then the histogram is built for each part to form the final LTP feature description of the original image. Therefore, a set of different thresholding numbers $\{\tau=0.1,0.2,0.5,1,2,\cdots,7\}$ are experimented, and it is found that $\tau=0.5$ yields optimal performance for this dataset.     

\begin{table}[ht]
	\centering
	\caption{Recognition rates as changing the threshold $t$ and the proposed approach order on extended Yale B dataset.} \label{table-1}
	\includegraphics[width=8.8cm,height=4cm]{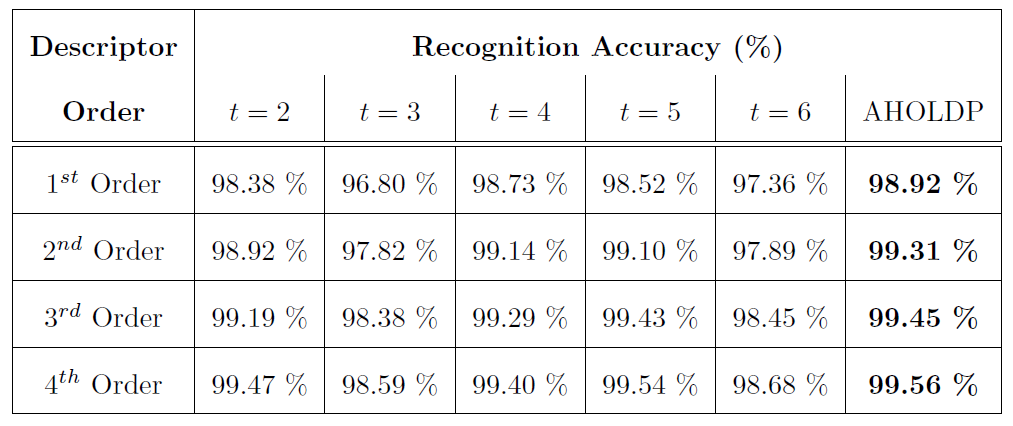}
\end{table}

\begin{table}[h!]
	\centering
	\caption{Performance comparison of the proposed methods with well-known face recognition algorithms on extended Yale B dataset.} \label{table-2}
	\includegraphics[width=8.8cm,height=2cm]{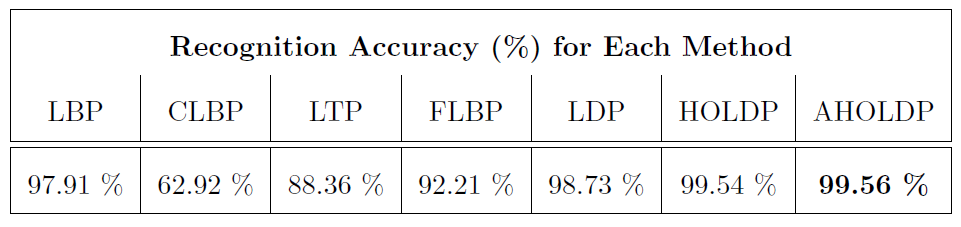}
\end{table}

\subsection{AT\&T Dataset (ORL)}
\begin{figure}[t!]
	\centering
	\hspace*{-.1cm}
	\includegraphics[width=.47\textwidth]{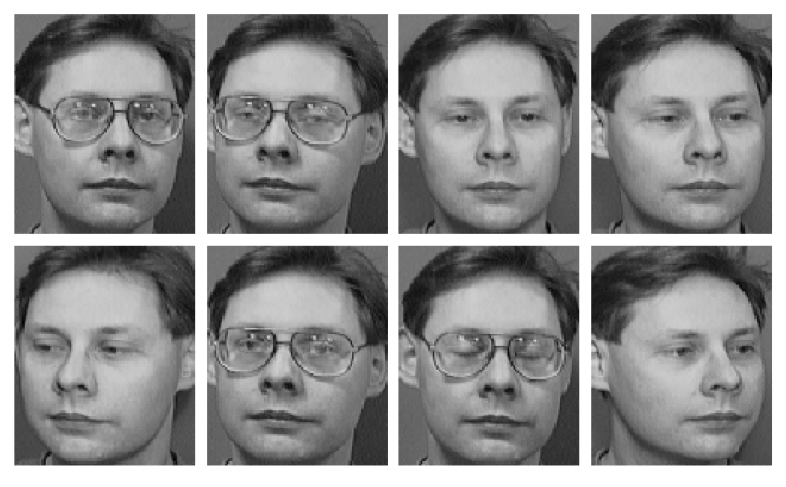}
	\caption{Samples of one subject from the ORL database.}
	\label{fig-8}       
\end{figure}

The ORL database contains $400$ face images corresponding to $40$ distinct subjects; each has $10$ different images. Some sample faces are shown in Fig. \ref{fig-8}. The images are taken at different times with different specifications, including varying slightly in illumination and pose, different facial expressions such as open and closed eyes, smiling and not smiling, and facial details like wearing glasses and not wearing glasses. The same procedure as in the previous section is applied, so we randomly selected half of the data for training ($5$ images/subject) and the other half for testing. Then we summarized the highest recognition rates as changing the order of the proposed descriptor in the range $(1-4)$ along with changing the threshold $t$ in the range $\{t=2,3,\cdots,6\}$ as well as with the adaptive HOLDP for each pixel of the input image in Table \ref{table-3}. Table \ref{table-4} presents the comparison results of LBP, CLBP, FLBP, LTP, and LDP with the proposed methods HOLDP and AHOLDP on ORL database. Note that we ran the same sets of training and testing for all methods, LBP, CLBP, FLBP, and LTP with a set of different thresholding numbers $\{\tau=0.1,0.2,0.5,1,2,\cdots,7\}$, and it is found that $\tau=2$ yields optimal performance for this database. 

\begin{table}[h]
	\centering
	\caption{Recognition rates as changing the threshold $t$ and the proposed approach order on ORL database.}
	\label{table-3}
	\includegraphics[width=8.8cm,height=4cm]{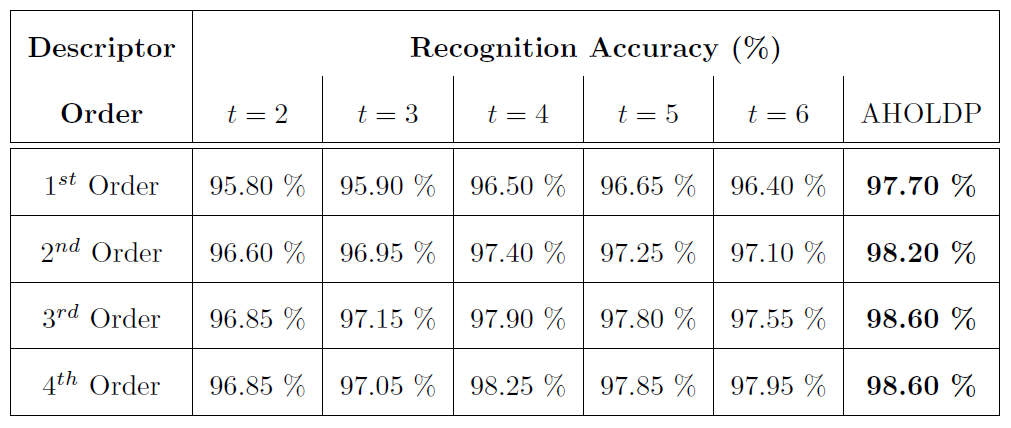}
\end{table}

\begin{table}[ht]
	\centering
	\caption{Performance Comparison of the Proposed methods with-well known face recognition algorithms on ORL database.}
	\label{table-4}
	\includegraphics[width=8.8cm,height=2cm]{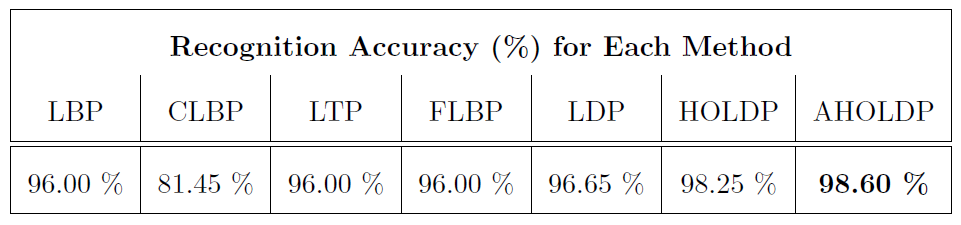}
\end{table}


\subsection{Georgia Tech (GT) Face Database}
\begin{figure}[t!]
	\centering
	\hspace*{-.1cm}
	\includegraphics[width=.48\textwidth]{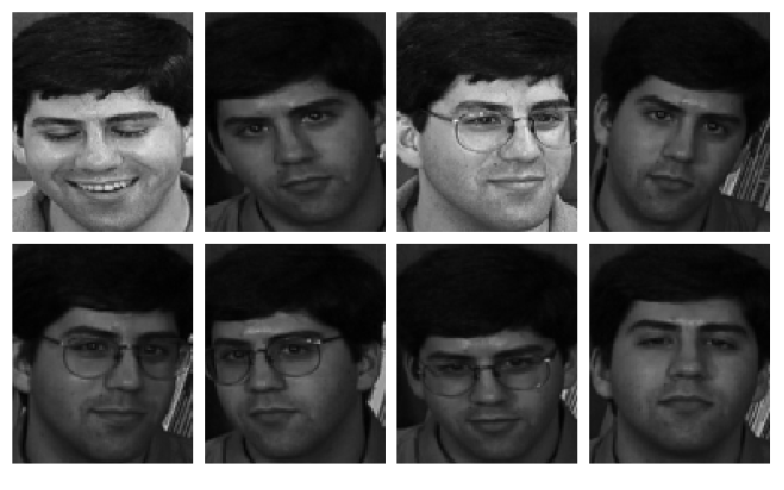}
	\caption{Samples of one subject from the GT database.}
	\label{fig-9}       
\end{figure}

The Georgia Tech (GT) face database consists $750$ color images corresponding to $15$ different images of $50$ distinct subjects. These images have large variations in both pose and expression and some illumination changes. Images are converted to gray scale and cropped into the size of $64\times64$. Some sample faces of one subject are shown in Fig. \ref{fig-9}. The same procedure as the previous sections is applied, so we randomly selected half of the data for training ($8$ images/subject) and the other half for testing. Then we summarized the highest recognition rates in Table \ref{table-5} to show the effectiveness of each neighborhood layer. Table \ref{table-6} presents the comparison results of LBP, CLBP, FLBP, LTP, and LDP with the proposed methods HOLDP and AHOLDP on GT database. Note that we ran the same sets of training and testing for all methods, LBP, CLBP, FLBP, and LTP with  LTP with a set of different thresholding numbers $\{\tau=0.1,0.2,0.5,1,2,\cdots,7\}$, and it is found that $\tau=3$ yields optimal performance for this dataset. 

\begin{table}[ht]
	\centering
	\caption{Recognition rates as changing the threshold $t$ and the proposed approach order on GT database.}
	\label{table-5}
	\includegraphics[width=8.8cm,height=4cm]{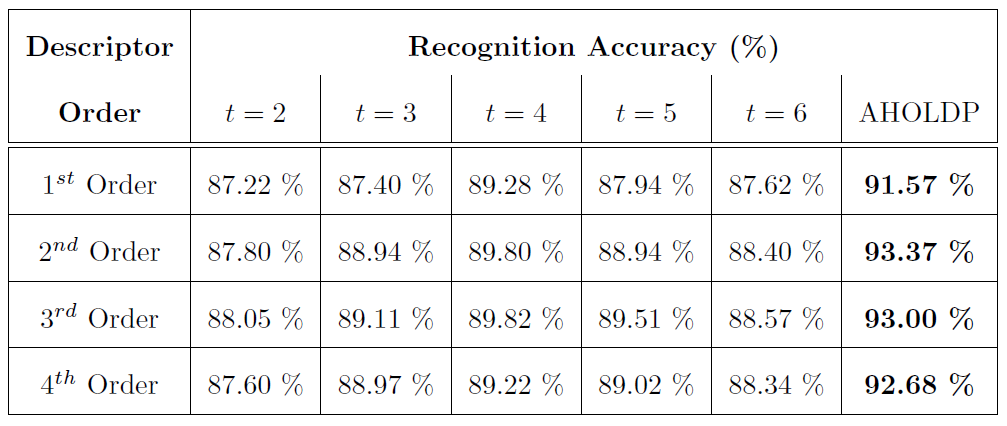}
\end{table}

\begin{table}[ht]
	\centering
	\caption{Performance comparison of the proposed methods with-well known face recognition algorithms on GT database.}
	\label{table-6}
	\includegraphics[width=8.8cm,height=2cm]{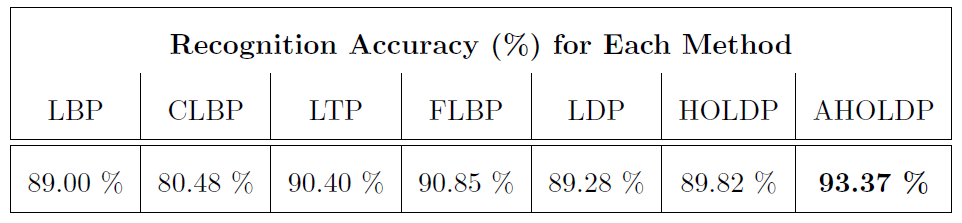}
\end{table}

\subsection{AR Database}
The AR face database contains over $4000$ color face images of 126 people, including frontal views of faces with different facial expressions, illumination conditions and occlusions. In our experiments, a subset with large variations in both illumination and expression was chosen, which corresponds to $50$ male subjects and $50$ female subjects. For each subject there are two sections, one for training and the other for testing. Each section contains $7$ images per subject. To show the effectiveness of each neighborhood layer, we summarized the highest recognition rates in Table \ref{table-7}. Table \ref{table-8} presents the comparison results of LBP, CLBP, FLBP, LTP, and LDP with the proposed methods HOLDP and AHOLDP on AR database. Note that we ran the same sets of training and testing for all methods, LBP, CLBP, FLBP, and LTP with  LTP with a set of different thresholding numbers of $\{\tau=0.1,0.2,0.5,1,2,\cdots,7\}$, and it is found that $\tau=5$ yields optimal performance for this database.

\begin{figure}[t!]
	\centering
	\hspace*{-.1cm}
	\includegraphics[width=.48\textwidth]{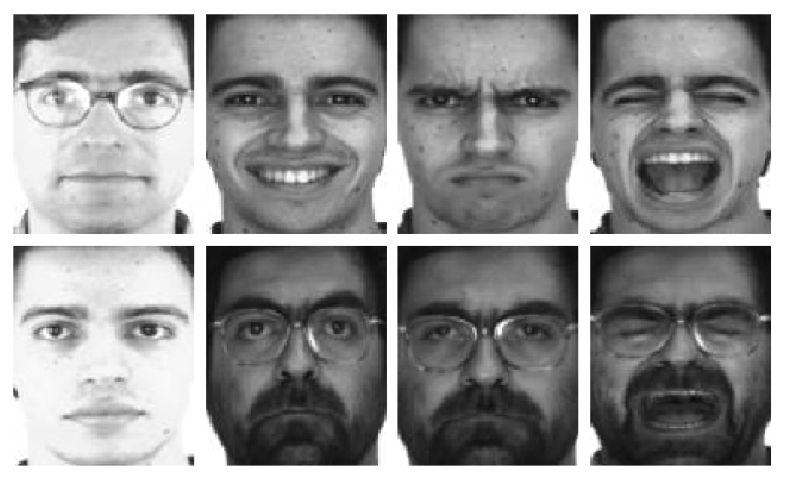}
	\caption{Samples of one subject from the AR database.}
	\label{fig-10}       
\end{figure}

\begin{table}[ht]
	\centering
	\caption{Recognition rates as changing the threshold $t$ and the proposed approach order on AR database.}
	\label{table-7}
	\includegraphics[width=8.8cm,height=4cm]{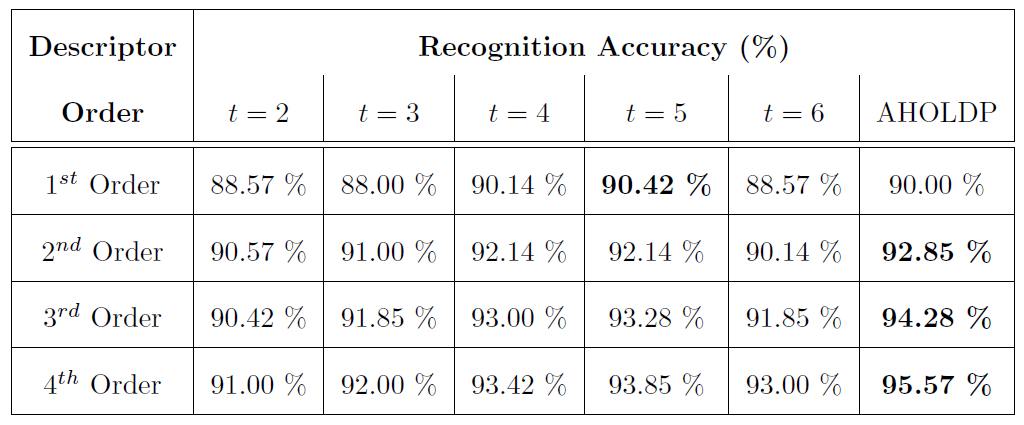}
\end{table}

\begin{table}[ht]
	\centering
	\caption{Performance comparison of the proposed methods with-well known face recognition algorithms on AR database.}
	\label{table-8}
	\includegraphics[width=8.8cm,height=2cm]{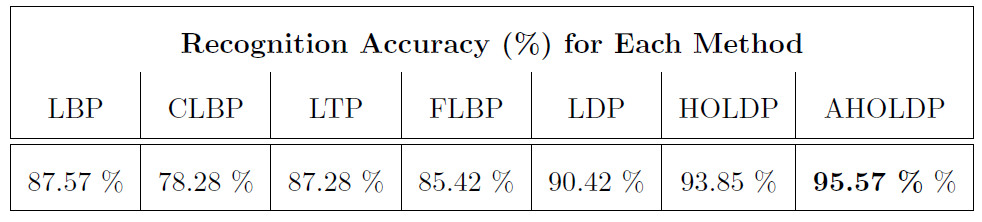}
\end{table}


\section{Discussion}
Derived from a general definition of texture in a local neighborhood, the conventional LDP encodes the directional information in the small $3\times3$ local neighborhood of a pixel, which may fail to extract detailed information, especially during changes in the input image due to illumination variations. To tackle this problem, a technique named HOLDP was developed. The key process of HOLDP is based on calculating the $n^{th}$ order directional variation patterns by encoding various distinctive spatial relationships from each neighborhood layer of a pixel in the pyramidal multi-structure way. The output of HOLDP provides a spatial histogram for modeling the distribution information of the input image. From the evaluation results, it has been found that the HOLDP algorithm can successfully perform feature extraction tasks and exceed a set of state-of-the-art descriptors in all test cases.

When it comes to the adaptiveness of the proposed HOLDP, the local directional patterns can be formed adaptively by comparing the $8$ neighboring pixels (excluding the central pixel) with their median of each neighborhood layer. If a neighboring pixel has a higher edge value than the median value (or the same value) then a $1$ is assigned to that pixel, which is otherwise a $0$. There are several different measures that have been experimented to adaptively find the relationship of the neighboring pixels of each layer such as thresholding the pixels with their mean, standard deviation, and variance, but we found that the median provides the best performance. In addition, from the experimental results (Tables 1, 3, 5, and 7) it can be observed that the highest accuracy rates of each descriptor order can be achieved when the threshold $t=4$ at the most test cases, which is the "middle" value in the list of the threshold $t$ that agrees with the median definition. 

Throughout the experimental results, we found that HOLDP and the adaptive HOLDP provide better performance for face recognition regardless of extreme variations of illumination environments and slight differences in pose and expression conditions. Furthermore, it is observed that a number of neighborhood layers and the threshold will affect recognition accuracy. From the results above it is clear that the high-order local patterns provide a stronger discriminative capability in describing detailed texture information than the first-order local pattern as used in the original LDP technique. When it comes to the adaptiveness of the proposed HOLDP, the adaptive HOLDP does provide the highest recognition rates than the HOLDP in all test cases. In general, considering all comparison results, we can assess that HOLDP can be a promising candidate for face recognition application.

\begin{figure}[!t]
	\centering
	\hspace*{-.2cm}
	\includegraphics[width=0.48\textwidth]{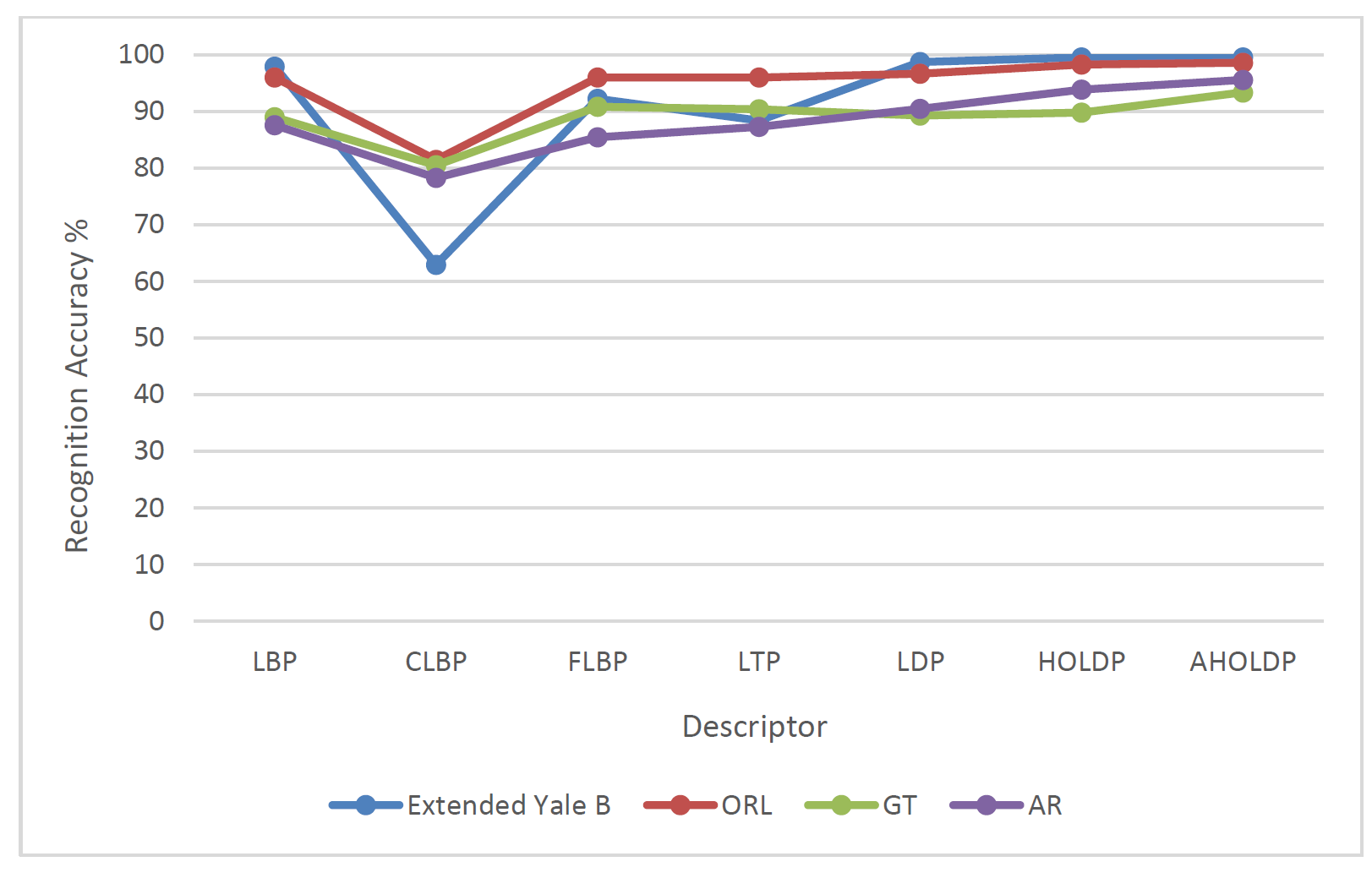}
	\caption{Recognition rates comparison on extended Yale B, ORL, GT, and AR databases.}
	\label{fig-11}       
\end{figure}  

\section{Conclusion}
In this paper, we have introduced a new local appearance feature extraction algorithm named HOLDP, which is capable of capturing discriminative information of still based face image. Throughout the performance evaluations, we found that HOLDP provides promising performance for face recognition regardless of extreme variations of illumination environments and slight differences in pose and expression conditions. In addition, compared to a set of state-of-the-art techniques, we conclude that our method provides better accuracy in all test cases. When it comes to the adaptiveness of the proposed HOLDP, the adaptive HOLDP does provide the highest recognition rates than the HOLDP in all test cases. In general, considering all comparison results, we can assess that the adaptive HOLDP can be a promising candidate for face recognition application. The work is progressing to explore the effectiveness of the proposed adaptive HOLDP on other applications such as texture classification, shape localization, deep learning applications, etc.

\bibliographystyle{IEEEtran}
\bibliography{REFERENCES}

\begin{thebibliography}{10}
\providecommand{\url}[1]{#1}
\csname url@samestyle\endcsname
\providecommand{\newblock}{\relax}
\providecommand{\bibinfo}[2]{#2}
\providecommand{\BIBentrySTDinterwordspacing}{\spaceskip=0pt\relax}
\providecommand{\BIBentryALTinterwordstretchfactor}{4}
\providecommand{\BIBentryALTinterwordspacing}{\spaceskip=\fontdimen2\font plus
\BIBentryALTinterwordstretchfactor\fontdimen3\font minus
  \fontdimen4\font\relax}
\providecommand{\BIBforeignlanguage}[2]{{%
\expandafter\ifx\csname l@#1\endcsname\relax
\typeout{** WARNING: IEEEtran.bst: No hyphenation pattern has been}%
\typeout{** loaded for the language `#1'. Using the pattern for}%
\typeout{** the default language instead.}%
\else
\language=\csname l@#1\endcsname
\fi
#2}}
\providecommand{\BIBdecl}{\relax}
\BIBdecl

\bibitem{1}
M.~Turk and A.~Pentland, ``Eigenfaces for recognition,'' \emph{Journal of
  cognitive neuroscience}, vol.~3, no.~1, pp. 71--86, 1991.

\bibitem{4}
T.~Ojala, M.~Pietik{\"a}inen, and D.~Harwood, ``A comparative study of texture
  measures with classification based on featured distributions,'' \emph{Pattern
  recognition}, vol.~29, no.~1, pp. 51--59, 1996.

\bibitem{6}
X.~Tan and B.~Triggs, ``Enhanced local texture feature sets for face
  recognition under difficult lighting conditions,'' in \emph{International
  Workshop on Analysis and Modeling of Faces and Gestures}.\hskip 1em plus
  0.5em minus 0.4em\relax Springer, 2007, pp. 168--182.

\bibitem{essa4}
A.~Essa and V.~Asari, ``Local edge/corner feature integration for illumination
  invariant face recognition,'' in \emph{The First International Conference on
  Applications and Systems of Visual Paradigms-VISUAL}, 2016, pp. 13--18.

\bibitem{5}
T.~Jabid, M.~H. Kabir, and O.~Chae, ``Local directional pattern (ldp) for face
  recognition,'' in \emph{2010 Digest of Technical Papers International
  Conference on Consumer Electronics (ICCE)}, 2010.

\bibitem{essa6}
A.~Essa and V.~Asari, ``Local boosted features for illumination invariant face
  recognition,'' \emph{International Conference on Electronic Imaging, Imaging
  and Multimedia Analytics in a Web and Mobile World 2017}, pp. 70--73, 2017.

\bibitem{essa5}
A.~Essa and K.~V. Asari, ``Fusing facial shape and appearance based features
  for robust face recognition,'' in \emph{2017 IEEE National Aerospace and
  Electronics Conference (NAECON)}.\hskip 1em plus 0.5em minus 0.4em\relax
  IEEE, 2017, pp. 7--10.

\bibitem{essa3}
A.~Essa and V.~K. Asari, ``Video-to-video pose and expression invariant face
  recognition using volumetric directional pattern,'' in \emph{VISAPP 2015 -
  Proceedings of the 10th International Conference on Computer Vision Theory
  and Applications, Volume 2, Berlin, Germany, 11-14 March, 2015}, 2015, pp.
  498--503.

\bibitem{essa7}
A.~Essa and K.~V. Asari, ``High order volumetric directional pattern for
  video-based face recognition,'' \emph{Mathematical Problems in Engineering},
  2019.

\bibitem{7}
T.~Ahonen, A.~Hadid, and M.~Pietik{\"a}inen, ``Face recognition with local
  binary patterns,'' in \emph{Computer vision-eccv 2004}.\hskip 1em plus 0.5em
  minus 0.4em\relax Springer, 2004, pp. 469--481.

\bibitem{8}
A.~Hadid, M.~Pietik{\"a}inen, and T.~Ahonen, ``A discriminative feature space
  for detecting and recognizing faces,'' in \emph{Computer Vision and Pattern
  Recognition, 2004. CVPR 2004. Proceedings of the 2004 IEEE Computer Society
  Conference on}, vol.~2.\hskip 1em plus 0.5em minus 0.4em\relax IEEE, 2004,
  pp. II--797.

\bibitem{9}
D.~Huijsman and N.~Sebe, ``Content-based indexing performance: A class size
  normalized precision,'' in \emph{Recall, Generality Evaluation, International
  Conference on Image Processing (ICIP'03)}, vol.~3, pp. 733--736.

\bibitem{10}
D.~Grangier and S.~Bengio, ``A discriminative kernel-based approach to rank
  images from text queries,'' \emph{Pattern Analysis and Machine Intelligence,
  IEEE Transactions on}, vol.~30, no.~8, pp. 1371--1384, 2008.

\bibitem{11}
A.~Oliver, X.~Llad{\'o}, J.~Freixenet, and J.~Mart{\'\i}, ``False positive
  reduction in mammographic mass detection using local binary patterns,'' in
  \emph{International Conference on Medical Image Computing and
  Computer-Assisted Intervention}.\hskip 1em plus 0.5em minus 0.4em\relax
  Springer, 2007, pp. 286--293.

\bibitem{12}
S.~Kluckner, G.~Pacher, H.~Grabner, H.~Bischof, and J.~Bauer, ``A 3d teacher
  for car detection in aerial images,'' in \emph{2007 IEEE 11th International
  Conference on Computer Vision}.\hskip 1em plus 0.5em minus 0.4em\relax IEEE,
  2007, pp. 1--8.

\bibitem{13}
T.~Ojala, M.~Pietikainen, and T.~Maenpaa, ``Multiresolution gray-scale and
  rotation invariant texture classification with local binary patterns,''
  \emph{IEEE Transactions on pattern analysis and machine intelligence},
  vol.~24, no.~7, pp. 971--987, 2002.

\bibitem{14}
T.~Ojala, M.~Pietik{\"a}inen, and T.~M{\"a}enp{\"a}{\"a}, ``A generalized local
  binary pattern operator for multiresolution gray scale and rotation invariant
  texture classification,'' in \emph{International Conference on Advances in
  Pattern Recognition}.\hskip 1em plus 0.5em minus 0.4em\relax Springer, 2001,
  pp. 399--408.

\bibitem{flbp}
D.~K. Iakovidis, E.~G. Keramidas, and D.~Maroulis, ``Fuzzy local binary
  patterns for ultrasound texture characterization,'' in \emph{International
  Conference Image Analysis and Recognition}.\hskip 1em plus 0.5em minus
  0.4em\relax Springer, 2008, pp. 750--759.

\bibitem{15}
T.~Ahonen and M.~Pietik{\"a}inen, ``Soft histograms for local binary
  patterns,'' in \emph{Proceedings of the Finnish signal processing symposium,
  FINSIG}, vol.~5, no.~9, 2007, p.~1.

\bibitem{clbp}
Z.~Guo, L.~Zhang, and D.~Zhang, ``A completed modeling of local binary pattern
  operator for texture classification,'' \emph{IEEE Transactions on Image
  Processing}, vol.~19, no.~6, pp. 1657--1663, 2010.

\bibitem{ltp}
X.~Tan and B.~Triggs, ``Enhanced local texture feature sets for face
  recognition under difficult lighting conditions,'' \emph{IEEE transactions on
  image processing}, vol.~19, no.~6, pp. 1635--1650, 2010.

\bibitem{Kirsch}
R.~A. Kirsch, ``Computer determination of the constituent structure of
  biological images,'' \emph{Computers and biomedical research}, vol.~4, no.~3,
  pp. 315--328, 1971.

\bibitem{23}
T.~Jabid, M.~H. Kabir, and O.~Chae, ``Robust facial expression recognition
  based on local directional pattern,'' \emph{ETRI journal}, vol.~32, no.~5,
  pp. 784--794, 2010.

\bibitem{24}
D.-J. Kim, S.-H. Lee, and M.-K. Sohn, ``Face recognition via local directional
  pattern,'' \emph{International Journal of Security and Its Applications},
  vol.~7, no.~2, pp. 191--200, 2013.

\bibitem{25}
F.~Zhong and J.~Zhang, ``Face recognition with enhanced local directional
  patterns,'' \emph{Neurocomputing}, vol. 119, pp. 375--384, 2013.

\bibitem{26}
A.~E. Essa and V.~K. Asari, ``Local directional pattern of phase congruency
  features for illumination invariant face recognition,'' in \emph{SPIE
  Defense+ Security}.\hskip 1em plus 0.5em minus 0.4em\relax International
  Society for Optics and Photonics, 2014, pp. 90\,940G--90\,940G.

\bibitem{essa1}
A.~Essa and V.~Asari, ``Face recognition based on modular histogram of oriented
  directional features,'' in \emph{Aerospace and Electronics Conference
  (NAECON) and Ohio Innovation Summit (OIS), 2016 IEEE National}.\hskip 1em
  plus 0.5em minus 0.4em\relax IEEE, 2016, pp. 49--53.

\bibitem{essa2}
A.~Essa and V.~K. Asari, ``Histogram of oriented directional features for
  robust face recognition,'' \emph{International Journal of Monitoring and
  Surveillance Technologies Research (IJMSTR)}, vol.~4, no.~3, pp. 35--51,
  2016.

\bibitem{yaleb1}
A.~S. Georghiades and P.~N. Belhumeur, ``Illumination cone models for faces
  recognition under variable lighting,'' in \emph{Proceedings of CVPR’98},
  1998.

\bibitem{yaleb2}
K.-C. Lee, J.~Ho, and D.~J. Kriegman, ``Acquiring linear subspaces for face
  recognition under variable lighting,'' \emph{Pattern Analysis and Machine
  Intelligence, IEEE Transactions on}, vol.~27, no.~5, pp. 684--698, 2005.

\bibitem{orl}
F.~S. Samaria and A.~C. Harter, ``Parameterisation of a stochastic model for
  human face identification,'' in \emph{Applications of Computer Vision, 1994.,
  Proceedings of the Second IEEE Workshop on}.\hskip 1em plus 0.5em minus
  0.4em\relax IEEE, 1994, pp. 138--142.

\bibitem{gt}
\BIBentryALTinterwordspacing
``Georgia tech face database.'' [Online]. Available:
  \url{http://www.anefian.com/face_reco.htm}
\BIBentrySTDinterwordspacing

\bibitem{ar1}
A.~M. Martinez, ``The ar face database,'' \emph{CVC technical report}, vol.~24,
  1998.

\bibitem{ar2}
A.~M. Mart{\'\i}nez, ``Recognizing imprecisely localized, partially occluded,
  and expression variant faces from a single sample per class,'' \emph{IEEE
  Transactions on Pattern analysis and machine intelligence}, vol.~24, no.~6,
  pp. 748--763, 2002.

\bibitem{libsvm}
C.-C. Chang and C.-J. Lin, ``Libsvm: a library for support vector machines,''
  \emph{ACM Transactions on Intelligent Systems and Technology (TIST)}, vol.~2,
  no.~3, p.~27, 2011.

\end{thebibliography}

\end{document}